\documentclass[10pt,twocolumn,letterpaper]{article}

\usepackage{iccv}
\usepackage{times}
\usepackage{epsfig}
\usepackage{graphicx}
\usepackage{amsmath}
\usepackage{amssymb}

\usepackage{gensymb}
\usepackage{mathtools}

\usepackage[pagebackref=true,breaklinks=true,letterpaper=true,colorlinks,bookmarks=false]{hyperref}

\iccvfinalcopy 

\def\iccvPaperID{****} 

\begin{document}

\title{Non-Lambertian Surface Shape and Reflectance Reconstruction \\Using Concentric Multi-Spectral Light Field }

\author{Mingyuan Zhou\\
University of Delaware\\
{\tt\small mzhou@udel.edu}
\and
Yu Ji\\
DGene, Inc\\
{\tt\small yu.ji@dgene.com}
\and
Yuqi Ding\\
Louisiana State University\\
{\tt\small yding18@lsu.edu}
\and
Jinwei Ye\\
Louisiana State University\\
{\tt\small jinweiye@lsu.edu}
\and
S. Susan Young\\
US Army Research Laboratory\\
{\tt\small shiqiong.s.young.civ@mail.mil}
\and
Jingyi Yu\\
ShanghaiTech University\\
{\tt\small yujingyi@shanghaitech.edu.cn}
}

\maketitle


\begin{abstract}
Recovering the shape and reflectance of non-Lambertian surfaces remains a challenging problem in computer vision since the view-dependent appearance invalidates traditional photo-consistency constraint. In this paper, we introduce a novel concentric multi-spectral light field (CMSLF) design that is able to recover the shape and reflectance of surfaces with arbitrary material in one shot. Our CMSLF system consists of an array of cameras arranged on concentric circles where each ring captures a specific spectrum. Coupled with a multi-spectral ring light, we are able to sample viewpoint and lighting variations in a single shot via spectral multiplexing. We further show that such concentric camera/light setting results in a unique pattern of specular changes across views that enables robust depth estimation. 
We formulate a physical-based reflectance model on CMSLF to estimate depth and multi-spectral reflectance map without imposing any surface prior. 
Extensive synthetic and real experiments show that our method outperforms state-of-the-art light field-based techniques, especially in non-Lambertian scenes.
\end{abstract}

\begin{figure}[t]
\begin{center}
\resizebox{1\linewidth}{!}{\includegraphics{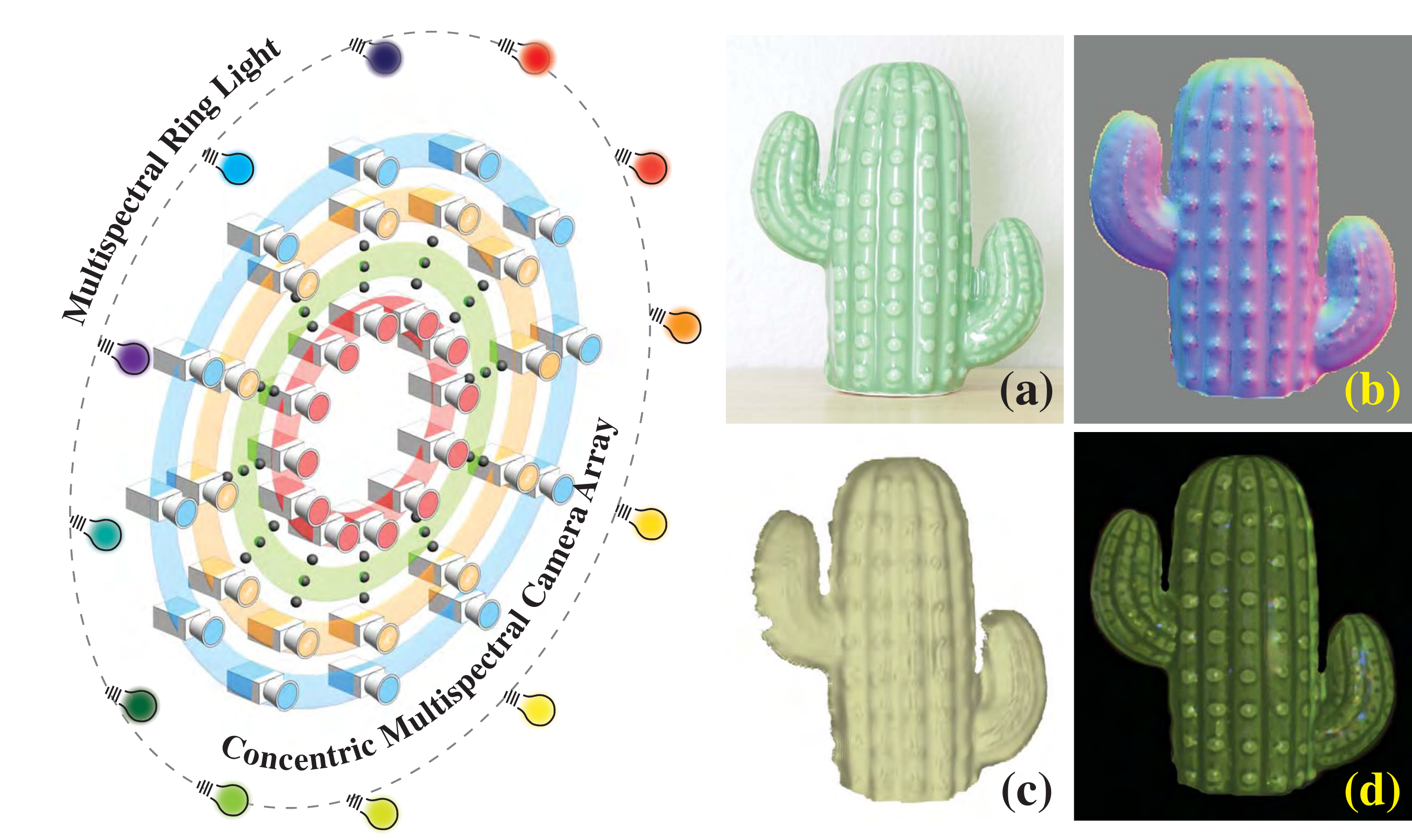}}
\end{center}
   \caption{Left: our concentric multi-spectral light field (CMSLF) acquisition system. We arrange cameras on concentric circles where each ring has the same number of cameras that capture at a specific spectrum. A multi-spectral ring light surrounds the cameras to provide direction-varying illumination for each camera ring. Right: our reconstruction results. (a) Photograph of the target object; (b) Recovered normal map; (c) Recovered 3D surface; and (d) Recovered reflectance map.}
\label{fig:hardware}
\end{figure}

\section{Introduction}
Surface shape and reflectance reconstruction from images is a fundamental problem in computer vision that can benefit numerous applications ranging from graphics rendering to scene understanding. Well established solutions based on multi-view stereo \cite{goesele2007multi,oxholm2014multiview,gotardo2015photogeometric} or photometric stereo \cite{vogiatzis2005using,goldman2010shape,fyffe2011single,barron2015shape} often assume Lambertian surfaces, from which light is equally reflected towards all directions. However, most real world objects have complex reflectance exhibiting view-dependent characteristics (such as specular highlights) that violates the Lambertian assumption and leads to the incorrect depth and reflectance estimation as a result.


In recent years, light field has emerged as a powerful tool in computer vision and graphics for 3D-related applications. A light field camera can be essentially viewed as a multi-view device. Notable examples include the hand-held light field camera \cite{ng2005light} and the light field camera array \cite{stdLF}: the former combines a lenticular lens array and a single high-resolution sensor with each lenslet emulating a pinhole camera while the latter uses multiple cameras that allows wider baseline and larger Field-of-View (FoV). Earlier uses of light field focused on refocused rendering \cite{ng2005light,levoy2006light} and view interpolation \cite{levoy1996light}. More recent approaches have employed light field for 3D reconstruction \cite{wanner2012globally,Yu_2013_ICCV,Jeon_2015_CVPR}. To handle non-Lambertian reflectance, focus cue \cite{tao2014depth}, angular coherence \cite{tao2015depth} and BRDF-invariants \cite{wang2016svbrdf,li2017robust} are proposed on light field data. However, additional surface priors such as smoothness or polynomial shape need to be imposed. 

In this paper, we introduce a novel concentric multi-spectral light field (CMSLF) design for recovering the shape and reflectance of surfaces with arbitrary material in one shot without imposing any surface prior. Our CMSLF acquisition system is shown in Fig.~\ref{fig:hardware}. We arrange cameras on concentric circles. Each ring has the same number of cameras that all capture images at a specific spectrum. In addition, we surround the concentric camera array with a multi-spectral ring light. Since we use narrowband spectral filters for cameras and light sources, we are able to simultaneously sample multiple viewpoints under varying lighting directions without interference via spectral multiplexing. We further show that under our concentric camera setting, the specular variation across views exhibits a unique pattern that helps separate specular components and enables robust depth estimation for non-Lambertian points. 

To estimate surface shape and reflectances, we formulate a dichromatic Phong reflectance model for CMSLF under the surface camera (S-Cam) representation \cite{yu2002scam}. An S-Cam models angular reflectance distribution with respect to a 3D scene point. It can be formed by tracing rays originated from the scene point back to the captured light field. By analyzing the reflectance model, we show that diffuse and specular surface points exhibit different characteristics under S-Cam. We use this property to initialize surface depth estimation and remove the specular components for all surface points. Finally, we jointly estimate the surface normal and multi-spectral reflectance coefficients from specular-free S-Cam and perform an iterative refinement. We conduct extensive synthetic and real experiments to demonstrate the accuracy and robustness of our approach. We also show that our method outperforms state-of-the-art light field-based method for shape and reflectance reconstruction, especially in non-Lambertian scenes.

\section{Related Work}
Our work is closely related to reflectance modeling and image-based surface shape and reflectance reconstruction.

Modeling surface reflectance is important to computer vision and graphics as it characterizes the surface material. The classical method in computer graphics uses the Lambertian model to characterize diffuse reflection and the Phong model to characterize specularity. Although this method is not theoretically correct, it is still widely used and indispensable in computer graphics due to its mathematical simplicity. To characterize complex surface reflectance, the bidirectional reflectance distribution function (BRDF) \cite{shafer1985using} that measures the ratio between incident irradiance and exit radiance at a surface point is commonly used. The full BRDF model of a surface requires huge parameter space as it exhausts all combinations of incident and exit lighting directions. A special case of the BRDF model is the dichromatic reflectance model, which was originally proposed by Shafer \cite{shafer1985using} to model dielectrics. It assumes that the BRDF of a surface can be decomposed into two additive components: the interface (specular) reflectance and the body (diffuse) reflectance. Since wavelength variations can be factorized from the two components, it is well suited for modeling multi-spectral reflectance. In our multi-spectral specular analysis, we combine the dichromatic reflectance model with the classical Phong model to characterize reflectance sampled by our angular light field.

Recovering surface geometry and reflectance from images is a fundamental problem in computer vision. The most popular two classes of methods are multiview photogrammetry \cite{jin2004shedding,jin2005multi,wu2011fusing,ghosh2011multiview,oxholm2014multiview, park2017robust,Fyffe2016NearInstantCO} and photometric stereo \cite{vogiatzis2005using,hernandez2007non,alldrin2008photometric,goldman2010shape,anderson2011color,fyffe2011single,barron2015shape}. The former recovers 3D shape by triangulating rays from multiple viewpoints while the latter performs reconstruction from a fixed viewpoint but under various lighting conditions. 
Although great success has been achieved on diffuse surfaces, specular highlights pose a challenge as they violate the color consistency assumption. Some methods \cite{tao2014depth,tao2016depth,lin2002diffuse} consider specular highlights as outliers and try to remove them. Some \cite{mallick2005beyond,sato1994temporal,nishino2001determining} rely on geometric and color distribution priors to compensate for specular regions. Recent work of Mecca \etal \cite{mecca2016unifying,tozza2016direct} separate specular from pure Lambertian reflection and treat them with different methods. However, their approach needs to take many (around ten) input images. Oxholm and Nishino \cite{oxholm2012shape} recover the shape and reflectance of the homogeneous surface from a single image captured under uncontrolled illumination. Zuo \etal \cite{zuo2017detailed} estimate surface geometry and albedo from RGB-D videos. Chandraker \cite{chandraker2014shape,chandraker2014camera} explores motion cue for recovering shape and reflectance of a homogeneous object under a single directional light source. Wang \etal \cite{wang2016svbrdf} extend similar motion cue to spatial-varying BRDF using light field. Li \etal \cite{li2017robust} improve the optimization framework for shape estimation with BRDF-invariants. 
In this work, we propose a novel concentric light field sampling scheme that results in unique specular variation pattern for robust depth estimation in non-Lambertian scenes.

\section{Concentric Multi-spectral Light Field} 




As shown in Fig.~\ref{fig:hardware}, our concentric multi-spectral light field (CMSLF) acquisition system is composed of multi-spectral cameras and light sources that arranged on coplanar and concentric circles. Each ring has the same number of cameras that are uniformly spaced and capture a unique spectrum. The surrounding multi-spectral ring light provides direction-varying illumination for each camera ring. It's worth noting that we use narrowband spectral filters for cameras and light sources. As a result, we are able to simultaneously sample multiple viewpoints under varying lighting directions without interference via spectral multiplexing.



To parameterize CMSLF, we adopt the classical two-plane parametrization (2PP) \cite{levoy1996light} light field representation. Since our cameras are on coplanar circles, we set the center-of-project (CoP) plane as the $st$ plane at $z=0$ and the image plane as the $uv$ plane at $z=1$. We use $st$ coordinate to index cameras and $uv$ coordinate to index pixels in the captured images. 

In our CMSLF, assume we have $m$ concentric camera rings in total and $n$ cameras on each ring, the camera or viewpoint position on the $st$ plane can be written as $(s(i,j),t(i,j)) = (r_j \cos{\phi_i},r_j \sin{\phi_i})$, where $i \in \{1,..., n\}$ is the camera index in each concentric ring; $j \in \{1,...,m\}$ is the ring index; $r_j$ is the radius of the $j$th ring; $\phi_i =(i-1)\tilde{\phi}$ is the spanned angle between the $i$th camera spoke and the x-axis ($\tilde{\phi} = 2\pi / n$ is the interval angle between neighboring camera spokes). The $j$th camera ring captures wavelength ${\lambda}_j$, $j \in \{1,...,m\}$.

On illumination side, since the lighting spectra match the ones that sampled by the camera array, the number of point light sources is equal to the number of rings (\ie, $m$). Assume the light source ring is on a circle with radius $r_l$, the position of the $j$th light source in 3D can be written as $P_j =  [r_l\cos{{\theta}_j},  r_l\sin{{\theta}_j}, 0]$ where ${\theta}_j = {\theta}_1 + (j-1)\tilde{\theta}$ (${\theta}_1$ is the angular position of the first light source and $\tilde{\theta} = 2\pi/m$ is the angular interval between neighboring light sources). 
We use the vector $ \mathbf{P} = [P_1;...;P_m]$ to represent the set of all lighting positions. The spectral filters used on the light sources are the same as the cameras $[{\lambda}_1;...;{\lambda}_m]$. Since we use narrowband spectral filters, the spectral illumination emitted from the $j$th point light source can only be received by the $j$th camera ring. 

\section{CMSLF Reflectance Model}
In this section, we formulate a reflectance model on CMSLF under the surface camera (S-Cam) representation \cite{yu2002scam}. By analyzing the reflectance model, we show that diffuse and specular surface points exhibit different characteristics under S-Cam. 

\subsection{Phong Dichromatic Model} \label{MSRM}
We adopt the Dichromatic Reflectance Model \cite{color1985} (DRM) for reflectance modeling. As DRM separates surface reflectance into body reflectance and interface reflectance and both terms account for geometry and color, DRM is suitable for modeling inhomogeneous materials.

Given a light source with the spectral distribution $E(\lambda)$ where $\lambda$ refers to wavelength, and a camera with spectral response function $Q(\lambda)$, the observed image intensity $I$ under DRM at pixel $p$ can be formulated as:

\begin{equation}
\label{eq:1}
\begin{split}
I(p)= &w_{d}(p)\int_{\lambda_1}^{\lambda_N}R(p, \lambda)E(p, \lambda)Q(\lambda)d\lambda\\
&+ w_{s}(p)\int_{\lambda_1}^{\lambda_N}E(p,\lambda)Q(\lambda)d\lambda
\end{split}
\end{equation}
where $[\lambda_1, \lambda_N]$ is the range of sampled wavelengths; $R(p,\lambda)$ is the surface reflectance; $w_{d}(p)$ and $w_{s}(p)$ are geometry-related scale factors. The first term in Eq.~\ref{eq:1} represents body reflectance that models light reflection after interacting with the surface reflectance. The second term represents interface reflectance that models light immediately reflected from the surface and thus causing specularites. 

We apply the numerical integration with step $\tilde{\lambda}$ on Eqn.~\ref{eq:1} with dropping pixel $p$ as:
\begin{equation}
\label{eq:2}
I=  w_{d}\mathbf{REQ} + w_{s}\mathbf{JEQ}
\end{equation}

where $\mathbf{J}$ is a row vector with all ones, $\mathbf{R} = [R(\lambda_{1}),R(\lambda_{1}+\tilde{\lambda}),...,R(\lambda_{N})]$,  $\mathbf{E} =diag(E(\lambda_{1}),E(\lambda_{1}+\tilde{\lambda}),...,E(\lambda_{N}))$, and $\mathbf{Q} = [Q(\lambda_{1}), Q(\lambda_{1}+\tilde{\lambda}),...,Q(\lambda_{N}) ]^T$.

To take scene geometry into consideration, we present the Phong dichromatic model that applies the classical Phong model and the near point lighting (NPL) model on top of the DRM (similar to \cite{tominaga2000estimating}). Specifically, the factors $w_{d}(p)$ and $w_{s}(p)$ are modelled in terms of lighting position, viewing direction, surface normal and roughness. The image intensity $I$ can be written as:

\begin{equation}
\label{eq:3}
I = \alpha\Big( \frac{ L\cdot N }{ \Vert P-X \Vert^2} \Big) \mathbf{REQ}  + \beta \Big( \frac{(D \cdot V)^{m}}{ \Vert P-X \Vert^2}  \Big)\mathbf{JEQ}
\end{equation}
where $N$ is the surface normal at a 3D point $X$; $P$ is the position of light source; $L = (P-X) / \Vert P-X \Vert $ is the normalized lighting direction; $V$ is the viewing vector; $D=2(L\cdot N)N-L$ is the reflection direction; $m$ is the shininess parameter that models the surface roughness; $\alpha$ and $\beta$ correspond to the diffuse and specular reflectivity of the surface. 


\subsection{Multi-spectral Surface Camera (MSS-Cam)}

Next, we apply the Phong dichromatic reflectance model on our CMSLF under Surface Camera (S-Cam) \cite{yu2002scam}. S-Cam characterizes the angular sampling characteristics of a light field from a 3D scene point. Given a 3D scene point, its S-Cam can be synthesized by tracing rays originated from the scene point into the light field to fetch color.

Applying S-Cam on our CMSLF, we obtain the multi-spectral S-Cam or MSS-Cam. We now derive intensities captured by the MSS-Cam using our reflectance model. Given a pixel $(u,v)$ in the virtual center view with camera position $(s,t) = (0,0)$, assume its corresponding 3D scene point is $X(u,v,z) = (x,y,z)$, we can synthesize its MSS-Cam $M_X$ from the captured multi-spectral light field images. Pixels in a column of $M_X$ are taken from cameras on the same ring that is sampled under a specific spectrum according to our concentric camera/light source arrangement. Each column captures the specular variation with respect to a single light source for non-Lambertian points. Pixels in the same row of $M_X$ are taken from cameras on different rings but along the same spoke. They, therefore, sample the spectral information. To obtain $M_X$, we trace rays from the point at $X$ to each camera in the CMSLF. For a pixel $(i,j)$ in our MSS-Cam, its sampling ray is from the camera at $(s(i,j),t(i,j))$. Therefore, applying Eqn.~\ref{eq:3}, we can write the intensity at a MSS-Cam pixel $M_X(i,j)$ as:

\begin{equation}
\label{eq:mx}
\begin{split}
  M_X(i,j) = &\alpha \Big( \frac{L_j\cdot N }{ \Vert P_j -X \Vert^2} \Big) c \mathbf{B_j E_j Q_j}  \\
  & \quad +  \beta \Big( \frac{(D_{j} \cdot V_{i,j})^{m}}{ \Vert P_j - X \Vert^2}  \Big)\mathbf{J E_j Q_j} 
\end{split}
\end{equation}
where $V_{i,j}$ is the viewing direction from the $X$ to the camera $(s(i,j),t(i,j))$; $D_{j}$ is the reflection direction of $L_j$; $c = [c_1,...,c_w]$ denotes the reflectance coefficient vector and $B_j$ is a $w \times k$ linear reflectance basis matrix under spectral range $[\lambda_j - \frac{(k-1)}{2}\tilde{\lambda}, \lambda_j + \frac{(k-1)}{2}\tilde{\lambda}]$, because the reflectance spectra $R$ can lie in a $w$-dimensional linear subspace \cite{parkkinen1989characteristic, maloney1986evaluation}. The $E_j$ and $Q_j$ are also under this spectral range and with size $k \times k$ and $k \times 1$ respectively. 

\begin{figure}[t]
\begin{center}
\resizebox{0.85\linewidth}{!}{\includegraphics{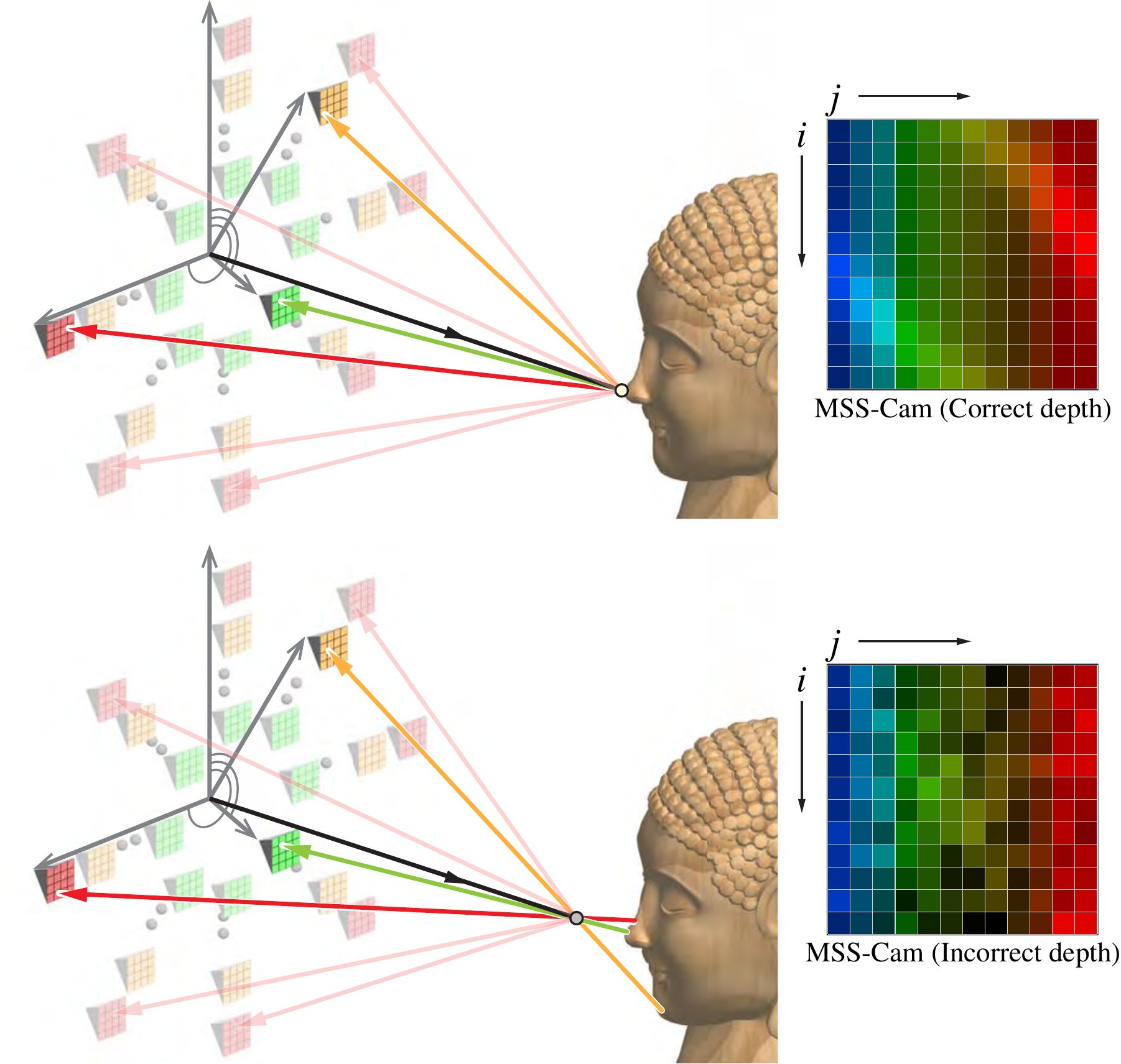}}
\end{center}
   \caption{Multi-spectral Surface Camera (MSS-Cam) sampling. Top: MSS-Cam sampled at the correct depth; Bottom: MSS-Cam for the same point but sampled at an incorrect depth.}
\label{fig:MSSCAM}
\end{figure}

\subsection{Diffuse vs. Specular Analysis}

\begin{equation}
\label{eq::fc}
   C (M_X)= \frac{1}{m} \sum_{j=1}^{m} std(M_X(1,j),...,M_X(n,j))
\end{equation}
where $std(\cdot)$ is the standard deviation. This function indicates that if taken pixels from the same column in an MSS-Cam, the standard deviation $C$ should be close to $0$ for diffuse points if the MSS-Cam $M_X$ is sampled at the correct depth. We therefore set a threshold on $C (M_X)$ to separate diffuse and specular points. 

\begin{figure}[t]
\begin{center}
\resizebox{1\linewidth}{!}{\includegraphics{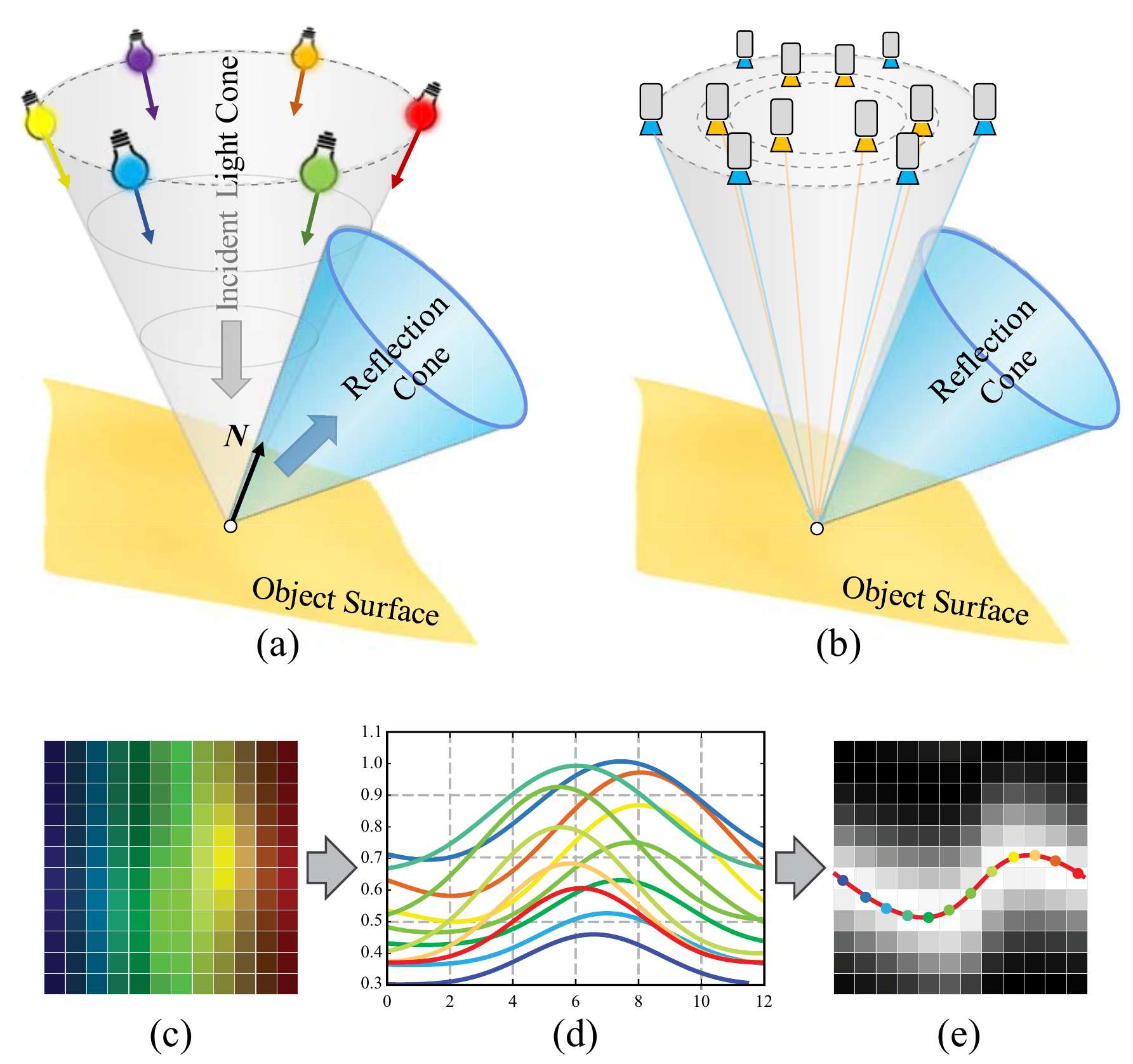}}
\end{center}
   \caption{The specularity variations in MSS-Cam exhibit unique pattern in our CMSLF. (a) The cone-shaped lighting directions results in a reflection cone that is symmetric to the normal. (b) Because of ring camera setting, the intensities from each column of the MSS-Cam will be changing on a periodic curve. (c) An MSS-Cam with specularity. (d) We plot the pixel intensities from the same MSS-Cam column and show that they for periodic curves. (e) The peaks of each curve in (d) form another periodic curve.  }
\label{fig:lightref}
\end{figure}

For specular points, we show that the specularity variation exhibits a unique pattern because of the concentric configuration of cameras/light sources. As shown in Fig.~\ref{fig:lightref}, the cone-shaped lighting directions results in the reflection directions also lying on a cone and the two cones are symmetric to the surface normal because of the reflection law. Since the light field camera sampling for each spectrum is on a circle, the intensities from each column of the MSS-Cam will be changing on a periodic curve with the camera angular angle $\phi$ from $0$ to $2\pi$ as shown in Fig.~\ref{fig:lightref}(d). Since Fourier series can be used to describe a periodic function, we approximate these periodic intensity variations on MSS-Cam by fitting Fourier series:
\begin{equation}
F(\phi) = a_0 + \sum_{p}a_p \cos{p\phi} + \sum_{p}b_p \sin{p\phi}    
\end{equation}

After we obtain the Fourier series models $F_{1},...,F_{m}$ for the $M_X$, the maximum intensity values on each curve form another periodic curve (see Fig.~\ref{fig:lightref} (e)). We fit a Fourier series $F_0$ to represent the curve. Therefore, if a point is specular, the follow consistency measurement should be satisfied: 

\begin{equation}
 S(M_X)= \frac{1}{m} \sum_{j=1}^{m} \Vert \mathbf{U}(j) - \mathbf{F}(j) \Vert +  \Vert \Phi - \mathbf{F}(0) \Vert 
\end{equation}
where $\mathbf{U}(j)= [ M_X(1,j),...,M_X(n,j)]$ are MSS-Cam pixels from the same column, $\mathbf{F}(j)= [ F_{j}(\phi_1),...,F_{j}(\phi_n)]$ are Fourier series fit for each column, $\Phi = [\phi_s^{(1)},...,\phi_s^{(m)}]$ are angles between viewing directions and reflection directions, and $\mathbf{F}(0) = [ F_{0}(1),...,F_{0}(m)]$ are Fourer series fit across the colums. 


\begin{figure}[t]
\begin{center}
\resizebox{1\linewidth}{!}{\includegraphics{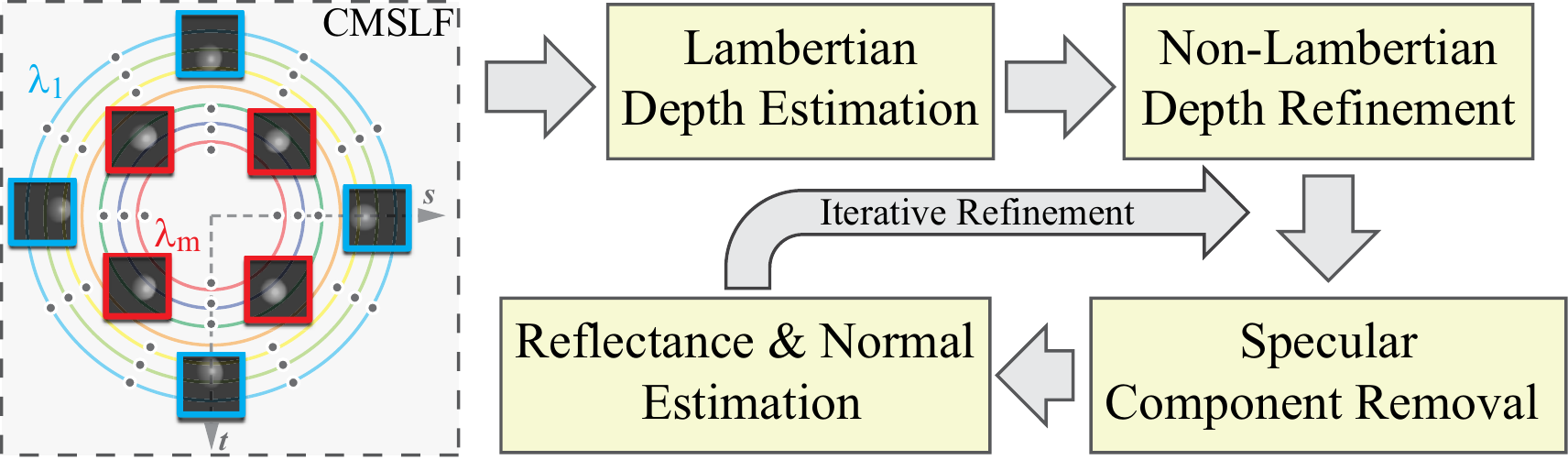}}
\end{center}
   \caption{Our shape and reflectance reconstruction pipeline.}
\label{fig:pipeline}
\end{figure}

\section{Shape and Reflectance Reconstruction}
Finally, we use above analysis on the MSS-Cam for surface shape and reflectance reconstruction. Our reconstruction pipeline is shown in Fig. \ref{fig:pipeline}.

\paragraph{Depth Initialization.} Given a pixel $(u,v)$ and its corresponding 3D point $X(u,v,z)$ in the virtual center view. We first apply our photo-consistency measurement on the MSS-Cams with every hypothetical depth $z$ of $X$ to initialize its depth as:

\begin{equation}
z^\prime = \operatorname*{argmin}_{z}  C(M_X)
\end{equation}
We classify this point as diffuse or specular point via the consistency measurement and a certain threshold. If it is the non-Lambertian point, we use our periodicity consistency to refine its depth as:


\begin{equation}
z^\prime = \operatorname*{argmin}_{z} S(M_X)
\end{equation}
Note that the Fourier series fitting is only desired by non-Lambertain points which are in the small regions, thus the time consuming for depth refinement is acceptable. Moreover, for any non-Lambertian point, given the estimated depth, we retrieve its MSS-Cam and then remove its specular components to obtain the specular-free MSS-Cam row vector $\mathbf{M}_X$. 

\paragraph{Specular Component Removal.}

For non-Lambertian points, we can exploit specular variations on their MSS-Cams to remove specularity. We first compute the vertical gradients of the MSS-Cam to remove the diffuse component in Eqn. \ref{eq:mx} as:

\begin{equation}
\begin{split}
\nabla & M_X(i,j)  = \Big( M_X(i+1,j) - M_X(i,j) \Big) \\
& = \beta((D_j \cdot V_{i+1,j})^{m} -(D_j \cdot V_{i,j})^{m})\frac{ \mathbf{J E_j Q_j}}{ \Vert P_j - X \Vert^2 }
\end{split}
\end{equation}
Then, we have: 
\begin{equation}
\begin{split}
 G_X(i,j)  = \nabla M_X(i,j) \frac{ \Vert P_j - X \Vert^2 }{ \mathbf{J E_j Q_j} }
\end{split}
\end{equation}

Given the pre-calibrated term $\mathbf{J E_j Q_j}$ and the lighting position $P_j$, we use the $\tilde{G}_X$ with our observed gradients to optimize the surface normal $N$, specular reflectivity $\beta$ and surface roughness $m$ simultaneously by:

\begin{equation}
\begin{split}
    \operatorname*{argmin}_{N, m, \beta} \sum_{i,j}  \Vert \tilde{G}_X(i,j) -  \beta ( (  D_j \cdot  V_{i+1,j} )^{m}  - ( D_j \cdot  V_{i,j} )^{m} ) \Vert
\end{split}
\end{equation}
where $D_j = 2(L_j \cdot N )N- L_j$. we apply the Levenberg-Marquardt method to solve all parameters and the solver will force the parameters to fit all specularity gradient variations. Given all specular parameters, we can remove the specular components on our MSS-Cam by:
\begin{equation}
\begin{split}
 A_X(i,j)  =  M_X(i,j) - \beta \Big( \frac{(D_{j} \cdot V_{i,j})^{m}}{ \Vert P_j - X \Vert^2}  \Big)\mathbf{J E_j Q_j} 
\end{split}
\end{equation}
Then we can generate a specular-free MSS-Cam row vector $\mathbf{A}_X = [ median(A_X(:,1)),...,median(A_X(:,m))]$ with the median values from each column of the specular-removal MSS-Cam.

\paragraph{Shape and Reflectance Estimation.} Finally, we introduce a multi-spectral photometric stereo method applying on the specular-free MSS-Cam to recover the surface normal and reflectance as:

\begin{equation}
    \operatorname*{argmin}_{N_X, c_X} (  (c_X  \mathbf{W}) \circ (\mathbf{L} N_X)^{T} -  \mathbf{M}_X)
\end{equation}

where $\circ$ is Hadamard product (element-wise multiplication), $\mathbf{L} = [L_1;...;L_m]$. The term $\mathbf{W}=  [\mathbf{W_1},...,\mathbf{W_m}]$ where $\mathbf{W_j} = \mathbf{B_jE_jQ_j}$ is pre-calibrated (see
supplementary materials for calibration details). Considering that the number of sampled spectra is greater than or equal to the dimensions of the reflectance spectra where $w \times 3 < m$, the above minimization can be formed into a linear least squares optimization, shown in supplemental material. When $w \times 3>m$, the linear least squares optimization can be tranfered to a over-determined bilinear optimization, we apply Levenberg-Marquardt algorithm to solve it.

After we obtain all recovered surface normals for all pixels in the center view, we update the depths from the estimated surface normals. Then, we re-retrieve all MSS-Cams with the updated depths and go through our pipeline one more time to refine the recovered surface normals. Given the final estimated reflectance coefficients, we can directly use recovered reflectance coefficients $c_X^{\prime}$ to get dense spectral reflectance of object by $\mathbf{R} = c_X^{\prime} \mathbf{B}$ where $\mathbf{B}$ is the dense spectral sampling reflectance basis. 


\section{Experiments}

We have validated our approach on both synthetic and real data. All experiments are conducted on a desktop with an Intel i7 7820 CPU (2.9GHz Quad-core) and 32G memory. Our algorithm is implemented in Matlab. All multi-spectral images are illustrated in RGB for better visualization.

\subsection{Synthetic Experiments} 

We conduct a multi-spectral renderer to render our multi-spectral light field from RGB image with a 3-dimensional reflectance linear basis. For the illumination, We use the spectral measures from the light sources in our real CMSLF setup.  

\begin{figure}[t]
\begin{center}
\resizebox{1\linewidth}{!}{\includegraphics{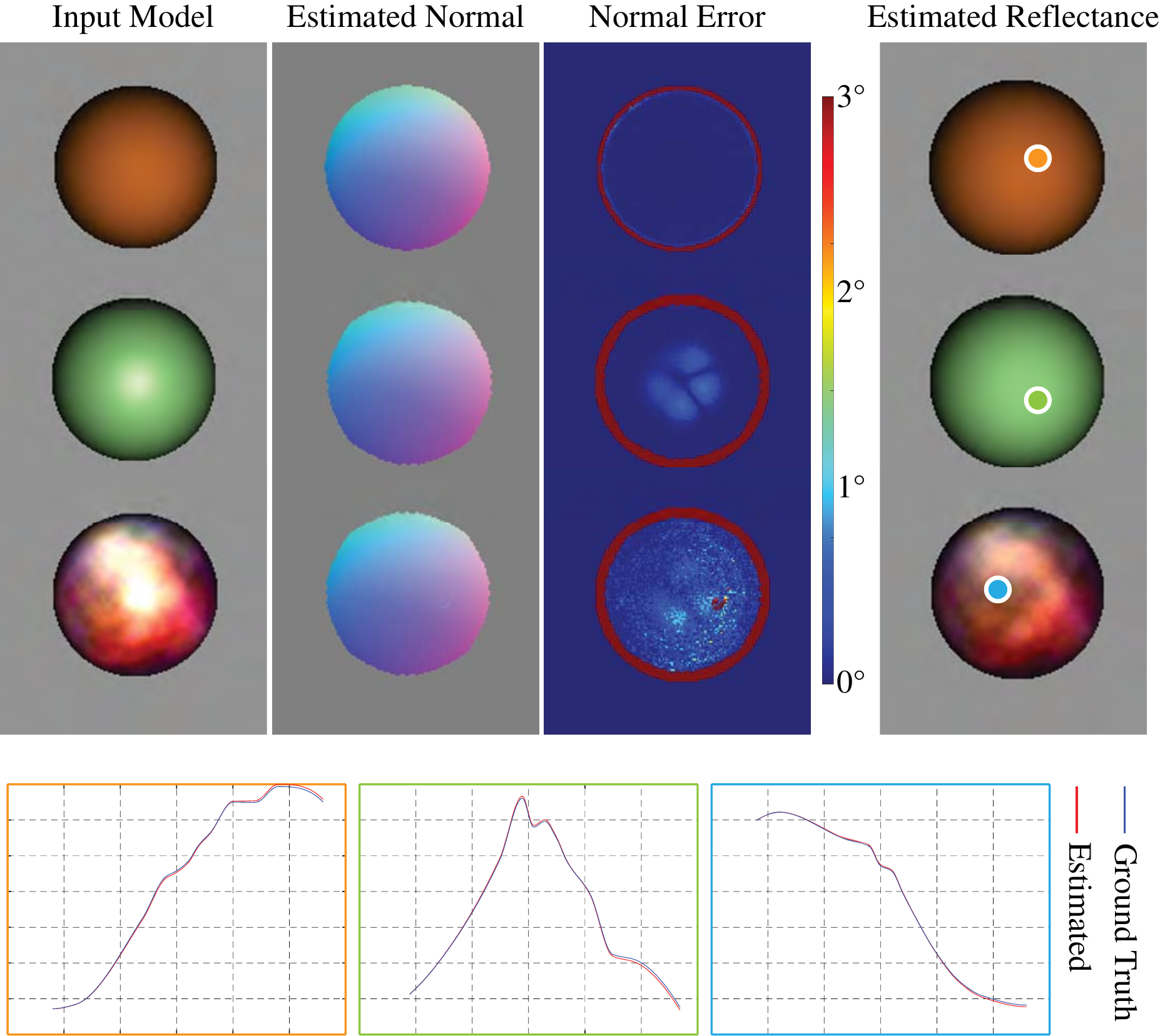}}
\end{center}
   \caption{ Qualitative  synthetic  results. The first column shows the input sphere with different reflectances. The second and third columns are our estimated normal maps and corresponding error maps. The last column is the estimated reflectance displaying in RGB, the dense recovered spectral reflectances compared to ground truth curves are presented at bottom. }
\label{fig:syn1}
\end{figure}

\begin{figure*}[htp]
\centering
\resizebox{1\linewidth}{!}{\includegraphics{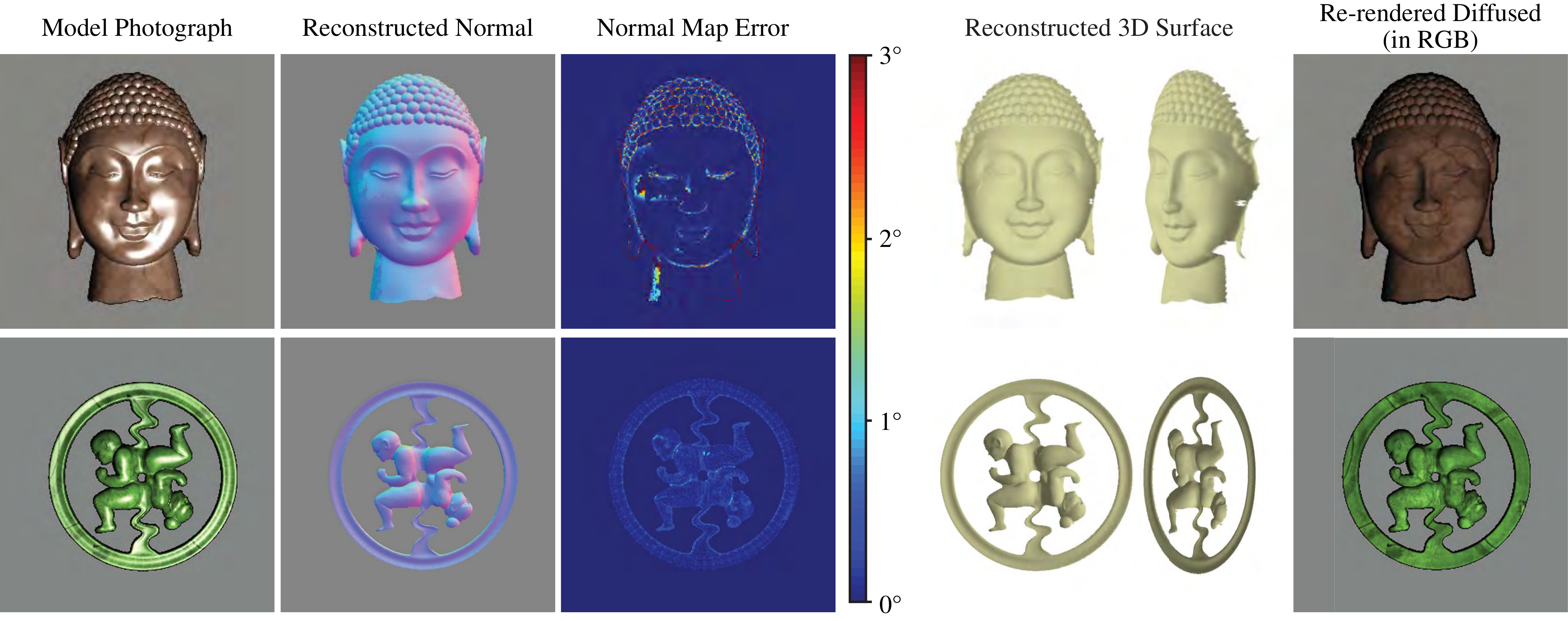}}
\caption{Shape and reflectance estimation on two complex synthetic scenes. The normal error maps and promising re-rendered diffused results, demonstrates that our aglorithm is robust against the specularity.}
\label{fig:syn2}
\end{figure*}

\begin{figure}
\begin{center}
\resizebox{1\linewidth}{!}{\includegraphics{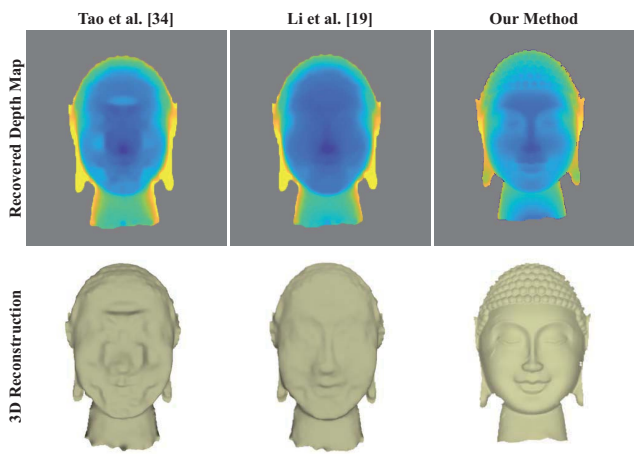}}
\end{center}
   \caption{Comparison with state-of-the-arts light field-based techniques. }
\label{fig:sync_compare}
\end{figure}

\begin{figure}
\begin{center}
\resizebox{1\linewidth}{!}{\includegraphics{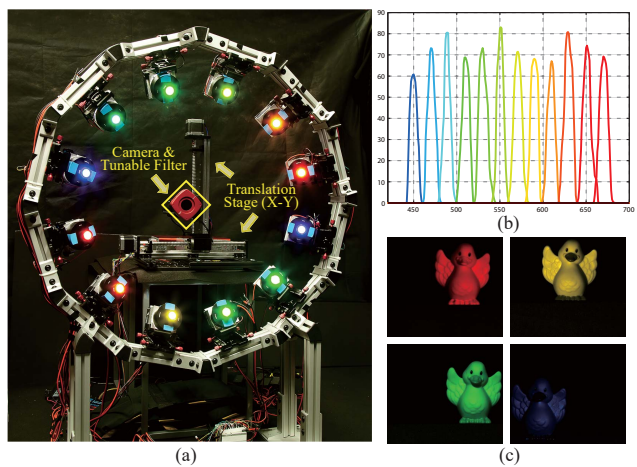}}
\end{center}
   \caption{(a) Our concentric multi-spectral light field (CMSLF) acquisition; (b) The illumination spectra; (c)Sample images from our captured CMSLF (We convert spectral images to RGB for better visualization). }
\label{fig:hardware_setup}
\end{figure}

\begin{figure*}
\centering
\resizebox{1\linewidth}{!}{\includegraphics{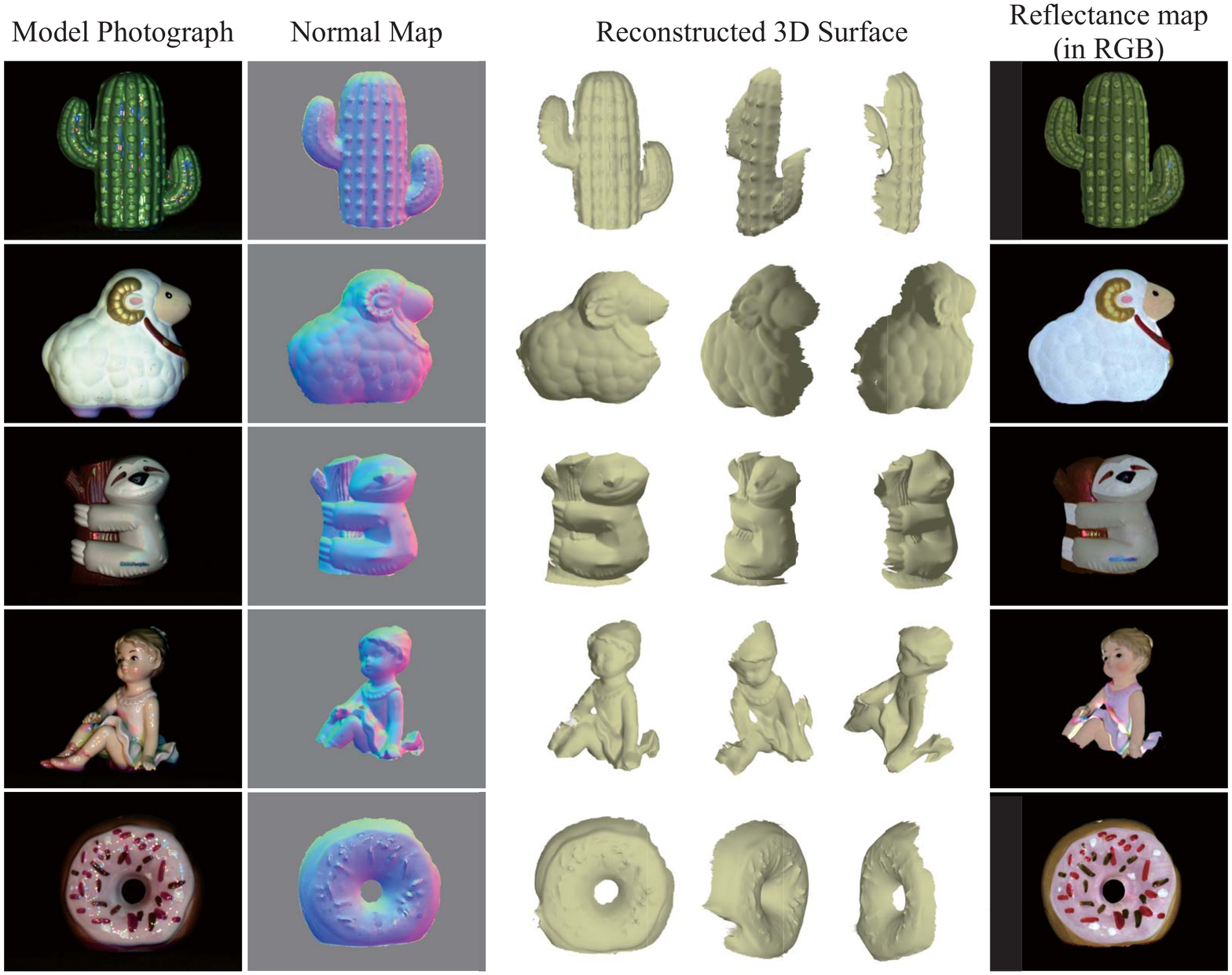}}
\caption{Shape and reflectance estimation results on real scenes with different materials. The images from the first column are captured by a RGB camera at camera position $(s(1,1),t(1,1))$ with all spectral light sources. The reconstructed shape are represented at the second and third columns. In order to visualize the recovered reflectance, we transfer the recovered dense spectral reflectance to the RGB relfectance shown in the last column. It can be seen that our approach can achieve favorable results.}
\label{fig:real_res}
\end{figure*}

We first test our scheme on a simple sphere for three different reflectances (uniform diffuse, specular and specular with texture). The diffuse coefficients are all set to 0.7 and specular coefficients for each material are respectively set to 0, 0.3, and 0.5. The roughness for the specular sphere is set to 8. In our experiment, we set the radius of the sphere to 20 and the distance between the CMSLF and sphere with 120. The synthetic image resolution is $320\times320$. The radius of our ring light source is 80 and contains 12 spectral light sources. We capture a $12\times 12$ concentric multi-spectral light field, the radius range of the concentric camera ring is from 29 to 40 with step 1. We set the filter wavelength from $440$ $nm$ to $660$ $nm$ at an interval of $20$ $nm$ for both cameras and light sources. For depth estimation, we discretize the depth from 108 to 125 with step 0.2, so that the sphere is modeled in terms of 76 depth layers. Fig. \ref{fig:syn1} shows the recovered surface normal and error map. we can see the all degree errors are less than 2$\degree$. The bottom row in Fig. \ref{fig:syn1} demonstrates examples of spectral reflectance estimation.

Next, we validate our approach on two complex scenes (buddha head and jadeware). Since these two models have more complicate geometric structures, We render the $12\times12$ multi-spectral light field with higher resolution ($500 \times 500$), we set the specular coefficient to 0.6 and 0.4 respectively, and the roughness to 20. The filter wavelengths are the same as the first sphere scene. The distances between objects and camera are 41 and 34. The radius of the camera ring is from 4 to 2.9 with step 0.1, and we discretize the depth ranges from 35 to 42 with step 0.1 for the buddha head and 33 to 35 with step 0.1 for the second, separately. Fig. \ref{fig:syn2} shows our reconstruction results. It can be seen that our algorithm can achieve reasonable spectral reflectance and 3D recovered geometries, and the maximum normal error is less than 3$\degree$ for both scenes. The artifacts around the right eye of the Buddha head is caused by extra-low spectral reflectance response over all wavelengths.

We also compare the reconstruction accuracy of our method with \cite{tao2016depth} and \cite{li2017robust}. We render the model in RGB images as their inputs with same resolution in a $12\times12$ light field arranged in 2D grid. From the Fig~. \ref{fig:sync_compare}, we can see that our method outperforms these two methods in recovering details. Our approach can robustly separate and remove specular components to achieve high frequency details of the surface. On the other hand, different lighting can benefit the specularity analysis in CMSLF.


\subsection{Real Experiments}

Finally, we construct a multi-spectral light field camera array to evaluate our algorithm on real-world data. 

To build the multi-spectral light field camera array, we mount a monochrome camera (Point Grey GS3-U3-51S5M-C) with 50mm lens on a translation stage to uniformly translate the camera position on a 2D plane (i.e., the $st$ plane), shown in Fig. \ref{fig:hardware_setup}. We mount a tunable liquid crystal spectral filter (KURIOS-WL1) in front of the camera to capture the scene under specified wavelengths. The camera resolution is 2448 $\times$ 2048 with 13 degree FoV. Our hardware setup is shown in Fig. \ref{fig:hardware}. To build the multi-spectral illumination, we mount twelve 30Watt LED chips onto a dodecagon frame, the distance between each LED chip and the center of the dodecagon is 50cm. We then place twelve narrow-band spectral filters range from 450 nm to 670 nm with step 20 nm in front of the LED chips. The distance between the acquisition system and the object is about 100cm. 

We calibrate the camera intrinsic parameters using traditional camera calibration method, and we also calibrate the position and spectral radiance of each light beforehand. The reflectance basis function and camera spectral response are pre-computed during the calibration. Details of our calibration process can be found in the supplemental material. 


Next, we test our algorithm with five objects with different reflectance (from diffuse to specular). Fig. \ref{fig:real_res} shows our reconstruction results. We can see that our method works well on real-world specular objects made of plastic, ceramics, etc. However, in presence of large occlusions and shadows, our method might fail to correct depths and reflectances as these points do not receive enough spectral samples. 

\section{Conclusion}
In summary, we have presented a concentric multi-spectral light field sampling scheme to recover shape and reflectance for non-Lambertain surfaces with arbitrary material. With our multi-spectral camera/light setting, we have sampled multiple viewpoints with varying lighting directions without interference. By using the different characteristics from diffuse and specular points on the MSS-Cam, we have proposed a new measurement to estimate depth and remove the specular components. Finally, we have reconstructed the surface normal and multi-spectral reflectance coefficients from the specular-free MSS-Cams and performed an iterative refinement. Comprehensive experiments show that our technique can achieve high accuracy and robustness in geometry and reflectance reconstruction. 



{\small
\bibliographystyle{ieee}
\bibliography{egbib}
}

\end{document}